\pdfoutput=1

\documentclass[11pt]{article}

\usepackage[final]{acl}

\usepackage{times}
\usepackage{latexsym}

\usepackage[T1]{fontenc}

\usepackage[utf8]{inputenc}

\usepackage{microtype}

\usepackage{inconsolata}

\usepackage{graphicx}

    \PassOptionsToPackage{numbers, compress}{natbib}
    
\usepackage[normalem]{ulem}
\usepackage{makecell} 
\usepackage{colortbl} 
\usepackage{xcolor}         
\usepackage{scalerel} 
\usepackage{amsmath}
\definecolor{uclablue}{rgb}{0.15, 0.45, 0.68}
\definecolor{custommagenta}{rgb}{0.1, 0.90, 1}

\usepackage{caption}  
\usepackage{enumitem} 
\setenumerate[1]{itemsep=0pt,partopsep=0pt,parsep=\parskip,topsep=5pt}
\setitemize[1]{itemsep=0pt,partopsep=0pt,parsep=\parskip,topsep=5pt}
\setdescription{itemsep=0pt,partopsep=0pt,parsep=\parskip,topsep=5pt}

\makeatletter
\g@addto@macro\normalsize{%
  \setlength{\abovedisplayskip}{6pt plus 2pt minus 2pt}%
  \setlength{\belowdisplayskip}{6pt plus 2pt minus 2pt}%
  \setlength{\abovedisplayshortskip}{4pt plus 2pt minus 2pt}%
  \setlength{\belowdisplayshortskip}{4pt plus 2pt minus 2pt}%
}

\usepackage{scalerel} 
\usepackage{booktabs} 
\usepackage{pdfpages}
\usepackage{multirow}
\usepackage{subfig}
\usepackage[capitalize]{cleveref}
\usepackage{xspace}
\usepackage{cleveref}
\usepackage{url}
\newcommand{\method}{\textsc{SpecVLM}\xspace}
\hypersetup{
    breaklinks,
    citecolor=uclablue,
    colorlinks=true,
    linkcolor=uclablue
}

\usepackage{pifont}

\makeatother

\title{
\scalerel*{\includegraphics{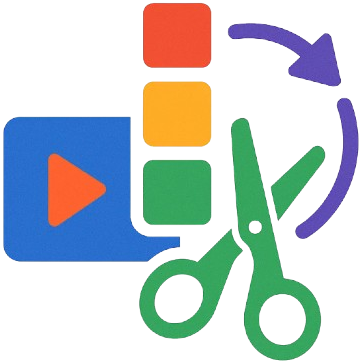}}{{\rule{2.2ex}{2.2ex}}}
\method: Enhancing Speculative Decoding of Video LLMs via Verifier-Guided Token Pruning}

\author{
  \textbf{Yicheng Ji\textsuperscript{1,2\ding{44}}}, 
  \textbf{Jun Zhang\textsuperscript{1,2\ding{44}}}, 
  \textbf{Heming Xia\textsuperscript{3}}, 
  \textbf{Jinpeng Chen\textsuperscript{4}},\\
  \textbf{Lidan Shou\textsuperscript{1,2},} 
  \textbf{Gang Chen\textsuperscript{1},} 
  \textbf{Huan Li\textsuperscript{1,2\ding{41}}}\\
  \textsuperscript{1}The State Key Laboratory of Blockchain and Data Security, Zhejiang University \\
  \textsuperscript{2}Hangzhou High-Tech Zone (Binjiang) Institute of Blockchain and Data Security \\
  \textsuperscript{3}Department of Computing, The Hong Kong Polytechnic University \\
  \textsuperscript{4}School of Computer Science, Beijing University of Posts and Telecommunications \\
  \small{\tt\{jiyicheng.cs, zj.cs, should, cg, lihuan.cs\}@zju.edu.cn, heming.xia@connect.polyu.hk, jpchen@bupt.edu.cn} 
}

\begin{document}
\maketitle

\let\oldthefootnote\thefootnote
\renewcommand{\thefootnote}{}

\footnotemark
\footnotetext{\textsuperscript{\ding{44}}Equal contribution. \textsuperscript{\ding{41}}Corresponding author.}

\let\thefootnote\oldthefootnote

\begin{abstract}
Video large language models (Vid-LLMs) have shown strong capabilities in understanding video content. However, their reliance on dense video token representations introduces substantial memory and computational overhead in both prefilling and decoding. To mitigate the information loss of recent video token reduction methods and accelerate the decoding stage of Vid-LLMs losslessly, we introduce \textsc{SpecVLM}\xspace, a \textit{training-free} speculative decoding (SD) framework tailored for Vid-LLMs that incorporates staged video token pruning.
Building on our novel finding that the draft model's speculation exhibits low sensitivity to video token pruning, \textsc{SpecVLM}\xspace prunes up to 90\% of video tokens to enable efficient speculation without sacrificing accuracy. To achieve this, we perform a two-stage pruning process: Stage I selects highly informative tokens guided by attention signals from the verifier (target model), while Stage II prunes the remaining redundant ones in a spatially uniform manner.
Extensive experiments on four video understanding benchmarks demonstrate the effectiveness and robustness of \textsc{SpecVLM}\xspace, which achieves up to 2.68$\times$ decoding speedup for LLaVA-OneVision-72B and 2.11$\times$ speedup for Qwen2.5-VL-32B.
Code is available at \url{https://github.com/zju-jiyicheng/SpecVLM}.



\end{abstract}

\section{Introduction}
Video large language models (Vid-LLMs)~\citep{li2024llava,ahmedqwen,lin2024video,fang2024mmbench} have demonstrated strong performance in video comprehension. Most Vid-LLMs use a sequential visual representation, encoding sampled frames into tens of thousands of video tokens alongside language prompts to achieve high generation performance. 
However, as long videos become more common, this design introduces significant memory and computational costs. For instance, LLaVA-OneVision~\citep{li2024llava} processes each video frame into 196 tokens, meaning a two-minute video at 60 FPS would require more than 1 million tokens by default without any reduction. 
The large number of video tokens increases the sequence length, resulting in quadratic attention overhead during prefilling. 
During decoding, the autoregressive nature of generation exacerbates the memory-bound issue, as the growing key-value (KV) cache must be loaded and stored in GPU memory alongside model parameters, limiting scalability and increasing latency~\citep{lin2024video}.

\begin{figure} [!t] 
\centering
    \includegraphics[scale=0.29]{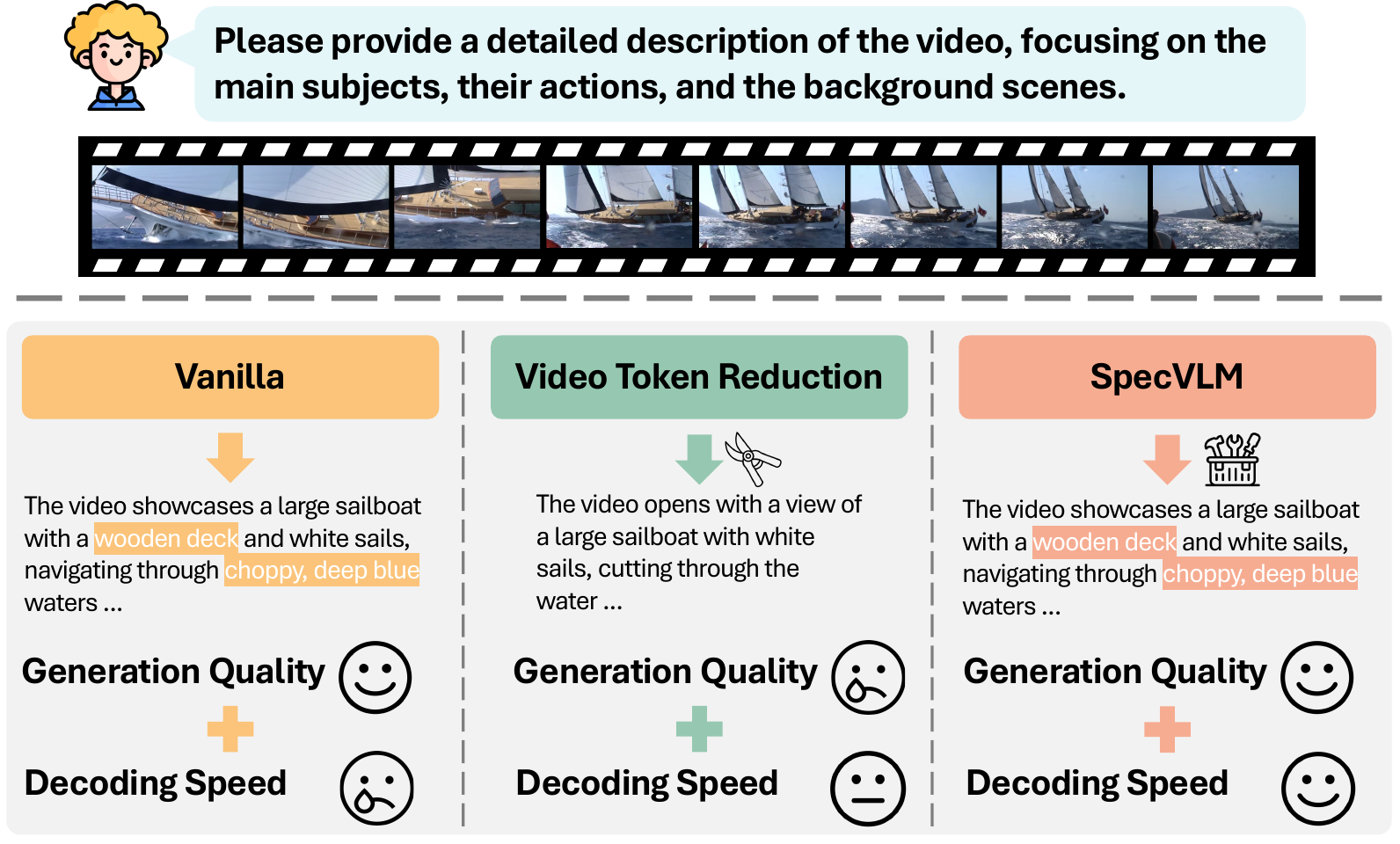}
    \caption{
    Comparison of vanilla autoregressive decoding, video token reduction, and our \method.
    }
    \label{intro_map}
\end{figure}

Recent studies have proposed token pruning strategies~\citep{chen2024image,liu2024multi,xing2024pyramiddrop,zhang2024sparsevlm,tao2024dycoke,shen2025fastvid,huang2024prunevid} to mitigate the rapidly growing overhead in storage, access, and computation incurred by tons of \textit{visual} tokens (including \textit{image} tokens and \textit{video} tokens). 
 These methods typically exploit token redundancy and importance variance, applying pruning during the prefilling stage to reduce subsequent memory and compute costs during decoding. However, by physically removing tokens from the input, they incur \textbf{inevitable information loss}—especially problematic in video understanding tasks where rich spatial and temporal cues are essential for maintaining generation quality. In addition, they offer \textbf{limited decoding speedup} due to repeated access to full model parameters at each generation step.
Fortunately, speculative decoding (SD) offers a promising solution to accelerate LLM decoding without sacrificing quality~\citep{leviathan2023fast} by using a lightweight \textbf{draft model} to propose multiple draft tokens, which are then verified in parallel by a \textbf{target model}.
However, deploying SD for Vid-LLMs is challenging.
First, autoregressive draft models for LLMs~\citep{du2024glide,li2024eagle,li2024eagle2} suffer from reduced efficiency in video scenarios because they require linearly increasing KV caches, which become the dominant bottleneck as the length of the video grows. 
Second, the visual context in video is relatively long and sparse, with significant redundancy. While existing SD methodologies tailored for long-context scenarios~\citep{sun2024triforce,chen2024magicdec,yang2025longspec} are modality-unaware, they fail to exploit the heavy redundancy and distinct attention patterns of video tokens (detailed in \cref{appendix:long-context sd}), leading to performance degradation (see \cref{ablation_study}).
These gaps motivate us to perform video token pruning for the draft model, thereby reducing its KV cache size and enhancing speculation efficiency.

Building on the observation that draft model speculation exhibits low sensitivity to random token pruning at low pruning ratios, we propose \method, a \textit{training-free} speculative decoding framework tailored for Vid-LLMs. As illustrated in~\cref{intro_map}, \method integrates staged video token pruning guided by verifier attention, extending decoding speed gains to high pruning ratios while preserving lossless generation quality.
Specifically, we utilize the attention guidance from the target model, and distinguish highly informative video tokens from redundant ones with low attention. Subsequently, we preserve the highly informative tokens following Top-P retention, while discarding the low attention tokens through spatially uniform reduction. 
By prefilling only the pruned video tokens, the draft model with memory-efficient KV caches achieves enhancing speculation.

To summarize, our main contributions are:

\begin{itemize}[itemsep=1.5pt, topsep=1.5pt, leftmargin=20pt]
    \item[(1)] To the best of our knowledge, we are the first to explore speculative decoding for lossless acceleration in Vid-LLMs, and further identify effective video token pruning as the silver bullet for the undesired slowdown in draft model speculation caused by video token explosion.
    \item[(2)] The surprising insensitivity of draft model speculation to random video token pruning at low pruning ratios sparks the emergence of \method, a training-free speculative decoding framework with verifier-guided staged video token pruning that pushes the performance boundary under aggressive pruning.
    \item[(3)] Thorough experiments on four video understanding benchmarks show \method prunes over 90\% of video tokens for the draft model while retaining nearly 90\% of the speculation accuracy, achieving up to 2.68$\times$ and 2.11$\times$ lossless decoding speedup for the LLaVA-OneVision and Qwen2.5-VL, respectively.
\end{itemize}

\begin{figure*} [t!] 
\centering
    \includegraphics[scale=0.4]{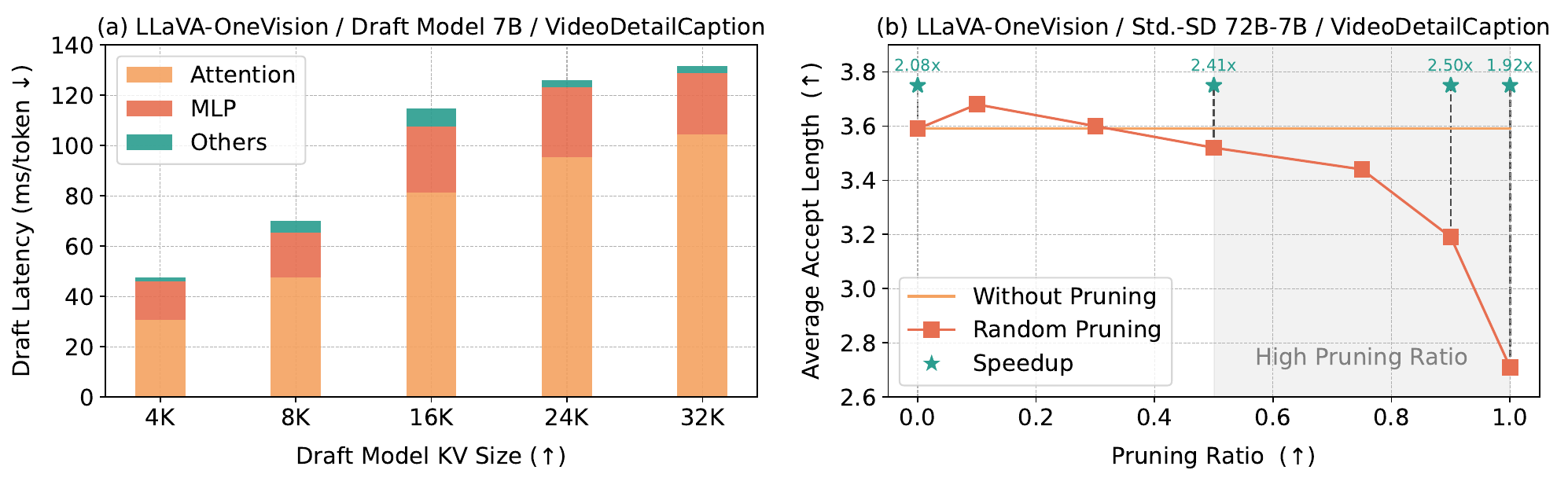}
    \caption{
    \textbf{(a)} Draft latency breakdown of LLaVA-OneVision-7B. Results are measured on a single NVIDIA A100 GPU by averaging the decoding time of 100 tokens. \textbf{(b)} Average accept length comparison on standard SD (Std.-SD). 
    }
    \label{bar_and_plot}
\end{figure*}

\section{Preliminary Study}
\subsection{Naive Speculative Decoding for Vid-LLMs}
\label{section_3.1}

Given a target model (verifier) $\mathcal{M}_{t}$ and a draft model $\mathcal{M}_{d}$, let $T_{t}$ and $T_{d}$ be the time for $\mathcal{M}_{t}$ and $\mathcal{M}_{d}$ to decode one token. For a predefined speculation length $\gamma$, $T_{t}^{\gamma}$ is the time for the target model to verify $\gamma$ tokens in parallel. Then, the time for each speculation decoding step is written as:
\begin{equation} \label{equation_1}
    T_{step}^{\gamma} = \gamma \cdot T_{d} + T_{t}^{\gamma}.
\end{equation}
Let $\tau$ be the average accept length of all decoding steps, the per token time of naive SD is given by:
\begin{equation} \label{equation_2}
    T_{token}^{\gamma} = {T_{step}^{\gamma}} ~/~ {\tau}.
\end{equation}
Hence, the 
speedup ratio is 
expressed as:
\begin{equation} \label{equation_3}
    \begin{aligned}
    \textit{Speedup} &= \frac{T_{t}}{T_{token}^{\gamma}} 
    = \frac{\tau \cdot T_{t}}{\gamma \cdot T_{d} + 
    T_{t}^{\gamma}} \\
    &= {\tau} ~/~ ({\gamma \cdot \frac{T_{d}}{T_{t}}
    + \frac{T_{t}^{\gamma}}{T_{t}}}).
    \end{aligned}
\end{equation}

\cref{equation_3} reveals that the speedup ratio is affected by (i) average accept length $\tau$, (ii) draft to target latency ratio $T_{d} / T_{t}$, and (iii) verification to target latency ratio $T_{t}^{\gamma} / T_{t}$, as described in previous study~\citep{chen2024magicdec}. For a normal batch size, $T_{t}^{\gamma} / T_{t}$ is close to 1 because the GPU parallelism is not fully utilized and the bottleneck still lies in memory bandwidth~\citep{patterson2005latency}. Therefore, the speedup ratio primarily depends on (i) and (ii).

Ideally, a draft model should have low latency, leading to a ratio $T_{d} / T_{t} \ll 1$. This condition is usually satisfied for LLMs in short-context scenarios by applying a parameter-efficient draft model. However, for Vid-LLMs with long video input, the latency bottleneck of the draft model shifts from the parameter scale to the accumulated KV cache, as illustrated in \cref{bar_and_plot} (a). As the input length grows, the expanding KV cache of the draft model leads to increased draft latency, especially in the attention layers where the entire KV cache is moved from GPU’s high-bandwidth memory (HBM) to its on-chip memory (SRAM) in each decoding step~\citep{sun2024triforce}. Hence, reducing the draft model KV size is a straightforward way to cut down draft latency and improve overall speedup for Vid-LLMs.

\subsection{Speculation Sensitivity for Token Pruning}
\label{section_3.2}
To reduce the draft model KV size, we incorporate video token pruning to shorten the video token sequence prefilled by the draft model. However, applying video token pruning may lead to a potential loss of visual information, which would in turn affect the accuracy of speculation. Concretely, as the pruning ratio increases, SD achieves faster speculation from the reduction in draft latency, but at the cost of a decreasing average accept length $\tau$ ( (i) in \cref{section_3.1}). 
To tackle this trade-off, we aim to address the following key question:
\textit{Is the draft model \textbf{sensitive} to video token pruning, i.e.,~can it maintain a stable accept length when part of the visual information is removed?}

To investigate this question, we intuitively apply \textit{random token pruning} for comparison with naive SD policy incorporating tree drafting. We use samples from the VideoDetailCaption~\citep{vdc} benchmark and employ LLaVA-OneVision to generate detailed captions of 256 tokens. We report the result in terms of average accept length $\tau$ as a direct measurement of speculation accuracy. 

 Our result in \cref{bar_and_plot} (b) indicates that random token pruning does not significantly compromise the average accept length under low pruning ratios ($\leq 50\%$), and even leads to improvements at certain pruning ratios. We attribute this to the excessive redundancy in long video input. Numerous redundant tokens divert attention away from important ones, and thus, moderate token removal may yield a positive impact. Moreover, when visual information is entirely removed (100\% pruning ratio), we observe a significant drop in the speculation accuracy and overall speedup, indicating that partial retention of video tokens is a better choice. This finding extends the previous work~\citep{gagrani2024speculative}, which suggests that the draft model may not require visual context as input. 

To summarize, our experiments demonstrate that \textbf{SD is not significantly sensitive to video token pruning under low pruning ratios}, which paves the way for its application. More precisely, it allows us to dramatically diminish the latency ratio $T_{d} / T_{t}$ without greatly affecting average accept length $\tau$ in \cref{equation_3}, serving as an enhanced solution.



Although random pruning can serve as a decent way to reduce the visual token number~\citep{wen2025stop}, it suffers from a considerable drop in the average accept length at high pruning ratios (i.e.,~$ >50\%$), as shown in \cref{bar_and_plot} (b). We argue that a more efficient and robust method is needed due to the following reasons: (i) As video length grows, methods under lower pruning ratios fail to effectively mitigate the high cost of KV cache. (ii) Random pruning is inherently stochastic and context-agnostic, which provides no deterministic lower bound on performance: with non-zero probability, it may discard all tokens corresponding to a critical semantic element (e.g.,~object boundaries or scene transitions in information-rich videos).
To address these issues, \cref{section_4} introduces a new video token pruning strategy that maintains a high accept length $\tau$ across diverse pruning ratios.

\begin{figure} [!t] 
\centering
    \includegraphics[scale=0.46]{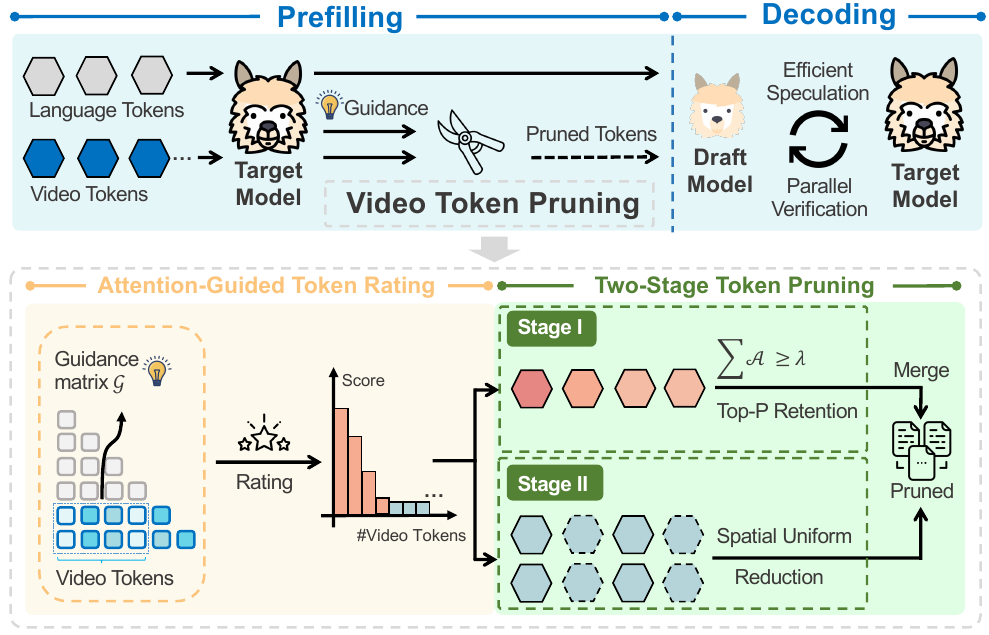}
    \caption{
    An overview of \method. In prefilling, the target model guides the pruning of video tokens for the draft model. In decoding, the draft model subsequently makes efficient speculation to accelerate the target model. The pruning process consists of two stages: Top-P retention of highly informative tokens and spatially uniform reduction of low-attention tokens.
    }
    \label{method_map}
\end{figure}

\section{\method: Enhancing Speculative Decoding for Vid-LLMs}
\label{section_4}
In this section, we introduce \method, an enhanced speculative decoding framework incorporated with video token pruning, as depicted in \cref{method_map}.
In \cref{section_4.1}, we study the attention pattern of vision-language input, and present our target model's attention-guided token rating scheme. In \cref{section_4.2}, we analyze the attention score distribution and propose a two-stage token pruning strategy accordingly.
Our method is simple yet effective, and can be applied in a plug-and-play manner.

\subsection{Attention-Guided Token Rating from Target Model}
\label{section_4.1}
To achieve accurate pruning, it is intuitive to preserve the important video tokens related to the query (e.g., the main subjects, actions, and background asked to describe). In previous studies, token importance has been evaluated using criteria such as \texttt{[CLS]} score~\citep{shang2024llava}, attention information from shallow layers~\citep{chen2024image}, or attention maps from small VLMs~\citep{shao2025growing,zhao2024stitch}. 
We argue that these methods differ from ours in their focus during token pruning: \textbf{while they aim to preserve output quality of the original model, our objective is to make the draft model’s output better aligned with that of the target model}.


\cref{method_map} illustrates how \method leverages attention signals from the verifier (target model) to guide video token pruning for the draft model. 
The resulting compact KV cache enables the draft model to perform efficient speculation, accelerating the target Vid-LLM. This design offers key advantages: (i) the target model, being more powerful and structurally similar to the draft, provides more accurate attention signals; (ii) allowing the draft model to ``see'' where the target model attends helps improve alignment; and (iii) from a latency perspective, the design adds minimal overhead, as target attention is computed regardless of speculation, and the draft only processes the pruned visual input.

\begin{figure} [t!] 
\centering
    \includegraphics[scale=0.28]{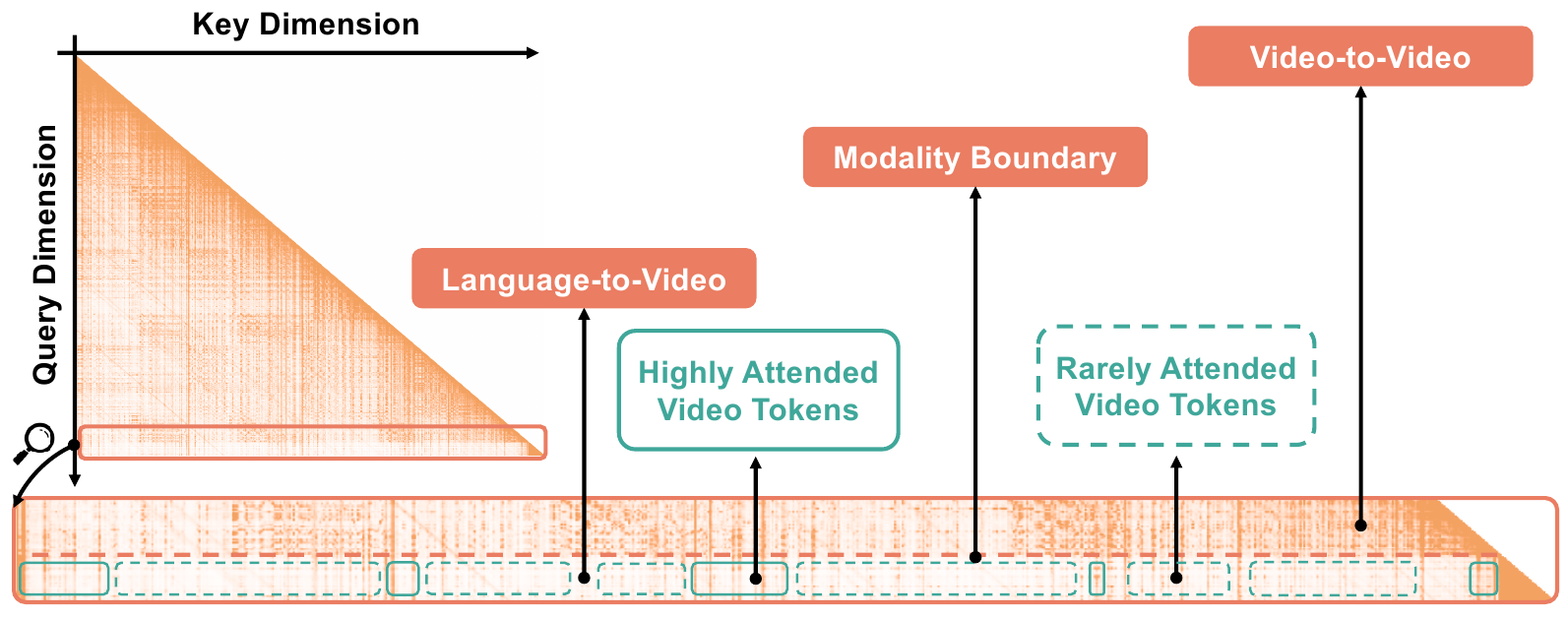}
    \caption{Attention map from LLaVA-OneVision-72B on an input comprising video tokens sampled from four frames in VideoDetailCaption and a language prompt.}
    \label{attention_map}
\end{figure}

To extract accurate attention signals from the target model, we follow previous work~\citep{tu2024vl,li2025mapsparse,li2025vistaenhancingvisiontextalignment} to study the attention pattern of vision-language input (see~\cref{attention_map}). By obtaining the full attention matrix, we observe a clear modality boundary emerging along the query dimension as~\citet{tu2024vl} described. In particular, video tokens attend to other video tokens densely while language tokens only pay attention to a few video tokens with high concentration, which reflects strong specificity.
Moreover, the tokens generated during decoding are language tokens below the modality boundary, where attention patterns exhibit similarity.
Based on the above observations, we extract \textbf{language-to-video attention scores} from the target model as guidance to rank video tokens for pruning.

Specifically, we take the query-dimensional part of the language modality $Q_{L}$ together with the key-dimensional part of the vision modality $K_{V}$, and the guidance matrix $\mathcal{G}$ is defined by:
\begin{equation} \label{equation_4}
    \begin{aligned}
    \mathcal{G} &= \textit{Attention}(Q_{L},K_{V}) 
    = M_{i,j},  \\ 
    & \text{and} \ (i,j) \in \{L,V\},
    \end{aligned}
\end{equation}
where $L$ and $V$ denote the language token set and video token set, respectively. Their cardinality are represented by $\|L\|$ and $\|V\|$. Next, we rate video tokens based on the average attention score they received in $\mathcal{G}$. The scores $\mathcal{A}$ are computed to guide token pruning, with formulation:
\begin{equation} \label{equation_5}
    \mathcal{A} = \{a_{j}\},\ \text{where} \ a_{j} = \frac{1}{\|L\|} \sum_{i=0}^{i < \|L\|} \mathcal{G}_{i,j}.
\end{equation}
This operation is performed by averaging across all layers and heads to obtain a holistic assessment.


\subsection{Two-Stage Token Pruning for Draft Model}
\label{section_4.2}
\begin{figure} [!t] 
\centering
    \includegraphics[scale=0.4]{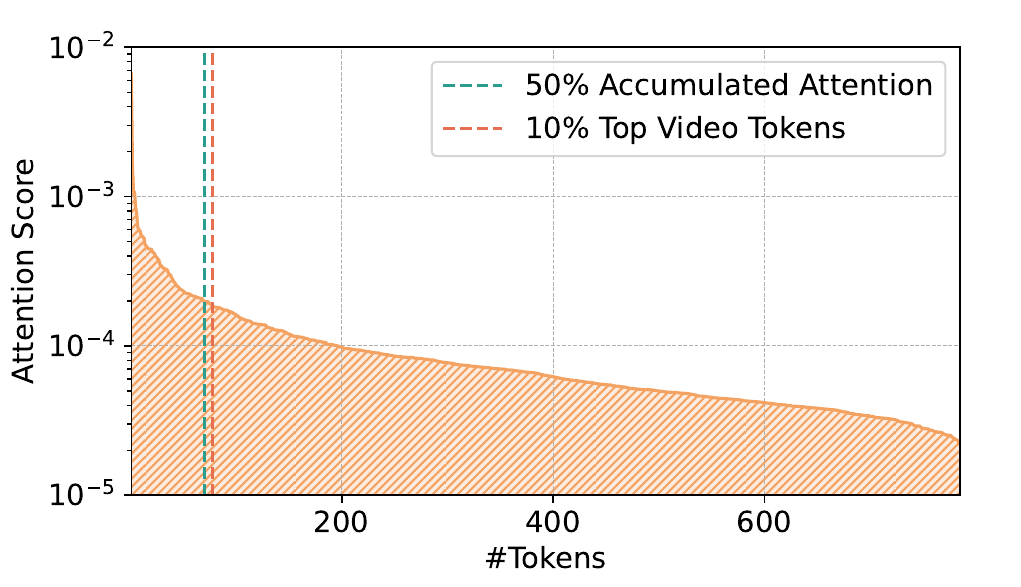}
    \caption{
    Long-tailed distribution of attention scores.
    }
    \label{longtail}
\end{figure}

Given an input video token set $V$ and a pruning ratio $r$, our goal is to prune redundant tokens and retain informative ones for the draft model, guided by the attention scores $\mathcal{A}$.
Notably, we observe that the attention scores in $\mathcal{A}$ follow a \textit{long-tailed distribution}, as shown in \cref{longtail}. This distribution reveals two key insights: (i) A minority of video tokens (10\%) accounts for of the total attention large percentage of attention scores (over 50\%), which makes their importance stand out compared to other tokens. (ii) The rest of the attention is amortized across the remaining 90\% of tokens,
making it difficult for the model to differentiate among them, as their attention scores are uniformly low.
We attribute this to the high redundancy among video tokens—most carry limited semantic value, yet collectively consume a significant share of attention due to their volume. This observation motivates a binary perspective in our pruning strategy.

\paragraph{Stage I: Top-P Retention of Highly Informative Tokens.}
To preserve highly influential video tokens, \method applies Top-P filtering to construct a candidate set whose cumulative attention scores exceed a threshold $\lambda_r$ ($\lambda_r \in [0,1]$), determined through a single offline evaluation on a small calibration set with no overlap with the test set~\footnote{Model-level selection of $\lambda_r$ is detailed in~\cref{appendix:implementation Details}.}.
This selection scheme enables dynamic adjustment of the token number for different queries, and ensures that sufficient visual information is retained. Concretely, for a given pruning ratio $r$, we first sort $\mathcal{A}$, and then repeatedly add tokens to the retention set $V_R$, until the proportion of their accumulated attention score reaches the predefined threshold $\lambda_r$:
\begin{equation} \label{equation_6}
\begin{aligned}
    V_R &= \arg\max_{c} \ {\mathcal{A}}, \  \text{where}\\
    c &= \min\left\{ c'\ \middle|\  \sum_{i=0}^{i < c'}a_{i} \geq  
    \lambda_{r}  \sum_{i=0}^{i < \|V\|}a_{i} \right\}.
\end{aligned}
\end{equation}


\paragraph{Stage II: Spatially Uniform Reduction of Low Attention Tokens.}
To handle tokens with uniformly low attention, we exploit the spatial redundancy of video context. 
As mentioned above, the subtle variations in attention scores make it difficult to distinguish between tokens in the ``tail'' part in \cref{longtail}. In \cref{ablation_study}, we empirically prove that continuing Top-P retention for these tokens would lead to suboptimal performance. Instead, we leverage the spatial continuity and strong local similarity inherent in video tokens, and select tokens to preserve at a fixed spatial interval $I$, where:
\begin{equation} \label{equation_7}
    I = \frac{\|V\|-\|V_{R}\|}{(1-r)\|V\|}.
\end{equation}

This operation is performed uniformly in space on the remaining token set $V \setminus V_{R}$.
Given that spatially adjacent video tokens are often highly similar, their partial removal incurs minimal visual information loss. Concurrently, retaining them at fixed spatial intervals allows the spatial structure to be effectively preserved. For a comprehensive study, we compare our spatially uniform reduction method with other temporal redundancy-based methods in \cref{ablation_study}, and find that spatial redundancy tends to be more significant in our SD setting. 
We analyze this phenomenon in \cref{ablation_study}, and choose to design our approach at the spatial level. 
Eventually, the tokens chosen in this step are collected as set $V_U$, which is merged with $V_R$ to form the final retention set of video tokens $V_R \cup V_U$. The draft model is then prefilled using video tokens in $V_R \cup V_U$ along with language prompts, with a KV cache reduced to $1-r$ of its original size and maximal preservation of video information.

\begin{table*}[!ht]
  \centering
  \scriptsize
  \setlength{\tabcolsep}{5pt} 
  \begin{tabular}{@{}l|l|ccc|ccc|ccc|ccc@{}}
    \toprule
    \multirow{2}{*}{Setup} & \multirow{2}{*}{Method} &
    \multicolumn{3}{c|}{VideoDetailedCaption} &
    \multicolumn{3}{c|}{MVBench} &
    \multicolumn{3}{c|}{MVLU} &
    \multicolumn{3}{c}{LongVideoBench} \\
    
    \cmidrule{3-5} \cmidrule{6-8} \cmidrule{9-11} 
    \cmidrule{12-14} 
    & & $\tau$ & Tokens/s & Speedup
    & $\tau$ & Tokens/s & Speedup 
    & $\tau$ & Tokens/s & Speedup 
    & $\tau$ & Tokens/s & Speedup \\
    \midrule
    \multirow{4}{*}{\makecell{Std.-SD \\ 72B-7B}} 
    & Vanilla & - & 2.94 & - & - & 2.93 & - & - & 3.03 & - & - & 2.75 & - \\
    & SD-Tree & 3.68 & 6.12 & 2.08$\times$ & 3.63 & 5.91 & 2.02$\times$ & 3.30 & 5.75 & 1.90$\times$ & 3.57 & 5.51 & 2.00$\times$ \\
    & SD-Rand & 3.19 & 7.36 & 2.50$\times$ & 3.32 & 6.75 & 2.30$\times$ & 2.83 & 6.03 & 1.99$\times$ & 3.14 & 6.73 & 2.45$\times$ \\
    & \method & \textbf{3.48} & \textbf{7.88} & \textbf{2.68$\times$} & \textbf{3.40} & \textbf{6.87} & \textbf{2.35$\times$} & \textbf{2.97} & \textbf{6.06} & \textbf{2.00$\times$} & \textbf{3.33} & \textbf{7.04} & \textbf{2.56$\times$} \\
    \midrule
    \multirow{4}{*}{\makecell{Self-SD \\ 7B-7B}} 
    & Vanilla & - & 13.31 & - & - & 12.36 & - & - & 11.82 & - & - & 13.55 & - \\
    & SD-Tree & 4.50 & 11.25 & 0.85$\times$ & 4.52 & 10.66 & 0.86$\times$ & 4.54 & 10.30 & 0.87$\times$ & 4.50 & 12.10 & 0.89$\times$ \\
    & SD-Rand & 3.81 & 15.73 & 1.18$\times$ & 3.93 & 15.99 & 1.29$\times$ & 3.64 & 13.97 & 1.18$\times$ & 3.59 & 16.83 & 1.24$\times$ \\
    & \method & \textbf{3.98} & \textbf{16.74} & \textbf{1.26$\times$} & \textbf{4.06} & \textbf{16.47} & \textbf{1.33$\times$} & \textbf{3.70} & \textbf{14.62} & \textbf{1.24$\times$} & \textbf{3.84} & \textbf{17.63} & \textbf{1.30$\times$} \\
    \bottomrule
  \end{tabular}
  \caption{Average accepted length $\tau$, decoding speed (tokens/s), and speedup of LLaVA-OneVision series on VideoDetailedCaption, MVBench, MVLU, and LongVideoBench. ``Vanilla'' refers to vanilla autoregressive decoding while ``SD-Tree'' denotes speculative decoding with draft trees. ``SD-Rand'' denotes SD-Tree incorporated with random token pruning. By default, $r=90\%$.}
  \label{table_1}
\end{table*}

\begin{table*}[!ht]
  \centering
  \scriptsize
  \setlength{\tabcolsep}{4pt} 
  \begin{tabular}{@{}l|l|ccc|ccc|ccc@{}}
    \toprule
    \multirow{2}{*}{Setup} & \multirow{2}{*}{Method} &
    \multicolumn{3}{c|}{VideoDetailedCaption} &
    \multicolumn{3}{c|}{MVBench} &
    \multicolumn{3}{c}{LongVideoBench} \\
    
    \cmidrule{3-5} \cmidrule{6-8} \cmidrule{9-11}
    & & $\tau$ & Tokens/s & Speedup
    & $\tau$ & Tokens/s & Speedup
    & $\tau$ & Tokens/s & Speedup \\
    
    \midrule
    \multirow{4}{*}{\makecell{Std.-SD \\ 32B-7B}} 
    & Vanilla & - & 4.88 & - & - & 5.50 & - & - & 4.91 & - \\
    & SD-Tree & 3.27 & 6.83 & 1.40$\times$ & 3.18 & 7.56 & 1.37$\times$ & 3.24 & 6.82 & 1.39$\times$ \\
    & SD-Rand & 3.21 & 9.93 & 2.03$\times$ & 3.17 & 10.13& 1.84$\times$ & 3.23 & 10.20 & 2.08$\times$ \\
    & \method & \textbf{3.23} & \textbf{9.99} & \textbf{2.05$\times$} & \textbf{3.18} & \textbf{10.17} & \textbf{1.85$\times$} & \textbf{3.28} & \textbf{10.35} & \textbf{2.11$\times$} \\
    
    \midrule
    
    \multirow{4}{*}{\makecell{Self-SD \\ 7B-7B}} 
    & Vanilla & - & 10.41 & - & - & 12.28 & - & - & 9.63 & - \\
    & SD-Tree & 4.38 & 8.65 & 0.83$\times$ & 4.31 & 10.02 & 0.82$\times$ & 4.17 & 7.82 & 0.81$\times$ \\
    & SD-Rand & 3.75 & 15.08 & 1.44$\times$ & 3.89 & 16.72 & 1.36$\times$ & 3.77 & 14.49 & 1.50$\times$ \\
    & \method & \textbf{3.83} & \textbf{15.59} & \textbf{1.50$\times$} & \textbf{3.92} & \textbf{16.80} & \textbf{1.37$\times$} & \textbf{3.84} & \textbf{15.33} & \textbf{1.59$\times$} \\
    \bottomrule
  \end{tabular}
  \caption{Average accepted length $\tau$, decoding speed, and speedup of Qwen2.5-VL series on VideoDetailedCaption, MVBench, and LongVideoBench (the sampling protocol of MVLU is incompatible). By default, $r=90\%$.}
\label{table_2}
\end{table*}

\paragraph{Bonus: Seamless Tree Attention Integration.} 
\method integrates tree attention by adopting the static tree structure from EAGLE~\citep{li2024eagle}, implemented via a specialized attention mask\footnote{Tree structure is detailed in \cref{Appendix:Tree}.}. The draft model generates multiple candidate tokens to form a draft tree, which are then verified in parallel by the target model. This design improves decoding speed by enabling more tokens to be accepted per forward pass.

\section{Experiments}

\label{set_up}
\paragraph{Target and Draft Models.}
We select two widely used Vid-LLM series: LLaVA-OneVision~\citep{li2024llava} and Qwen2.5-VL~\citep{ahmedqwen}. Two representative SD settings are employed. (i) \textbf{Standard Speculative Decoding (Std.-SD)}: using a smaller Vid-LLM from the same model series as the draft model. (ii) \textbf{Self-Speculative Decoding (Self-SD)}: using the model with original parameters and pruned KV cache as draft model.
This setting avoids introducing additional models, allowing the acceleration gains to come solely from video token pruning, which can serve as a more general solution.
We do not introduce a training process for the draft model as the major bottleneck is KV caches rather than model parameters, which aligns with~\citet{chen2024magicdec,yang2025longspec}.


\paragraph{Tasks and Benchmarks.}
Since \method mainly focuses on the acceleration during decoding, we evaluate its performance in video captioning and video description tasks, which require generating long text paragraphs. The benchmarks we use include VideoDetailCaption~\citep{vdc}, MVBench~\citep{li2024mvbench}, MVLU~\citep{zhou2024mlvu}, and LongVideoBench~\citep{wu2024longvideobench}. Experimental details are elaborated in \cref{appendix:implementation Details}.

\paragraph{Metrics.}
We assess the acceleration effects using: (i) decoding speed and wall time speedup ratio relative to vanilla autoregressive decoding, and (ii) average accept length $\tau$, which directly reflects the speculation accuracy. Output quality is not evaluated since our method is lossless.

\begin{figure*} [!t]
\centering
    \includegraphics[scale=0.4]{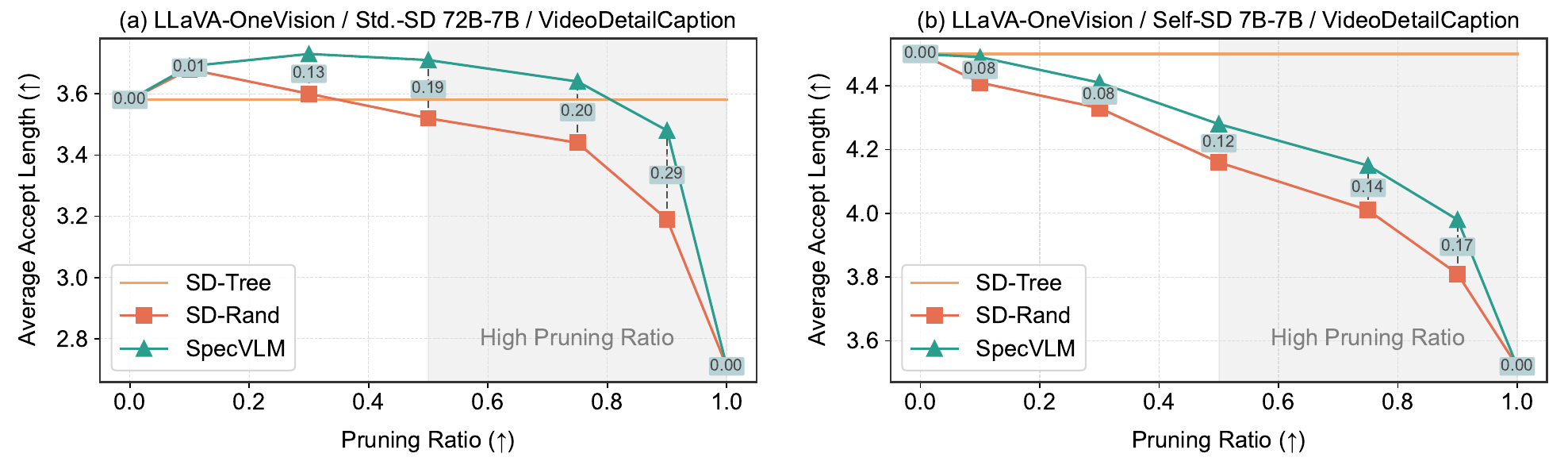}
    \caption{Average accepted length across different baselines as pruning ratios scale up.
    }
    \label{llava_saling_comparison}
\end{figure*}

\begin{figure*} [!t]
\centering
    \includegraphics[scale=0.4]{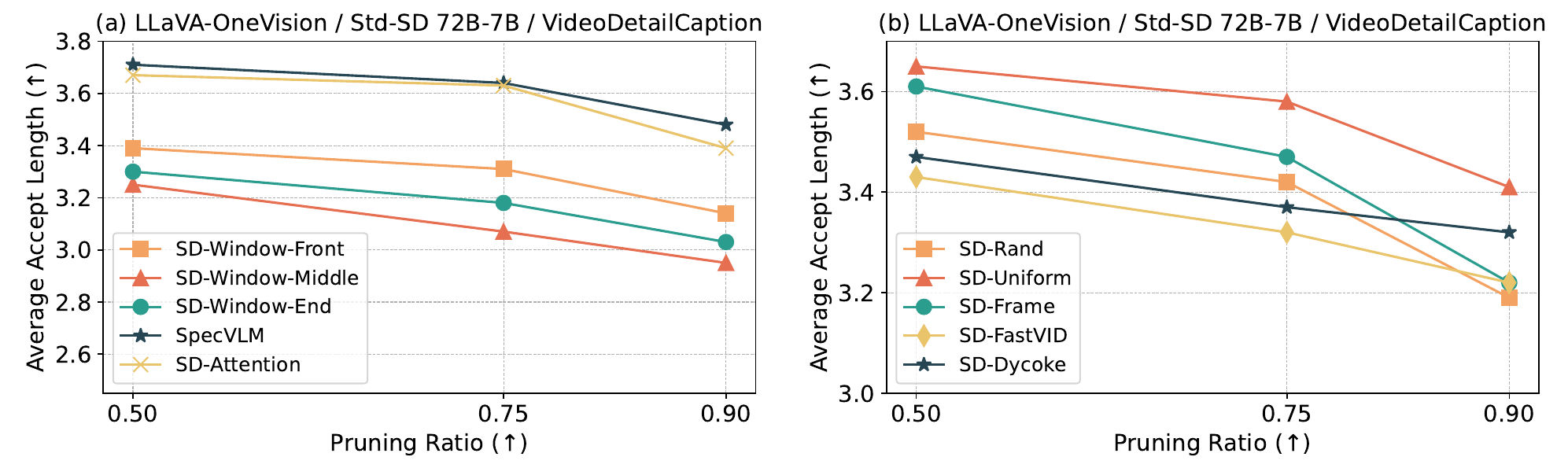}
    \caption{\textbf{(a)} \method vs. window-based methods. \textbf{(b)} Spatial vs. temporal redundancy-based methods.}
    \label{llava_baseline}
\end{figure*}

\begin{figure*} [!t]
\centering
    \includegraphics[scale=0.4]{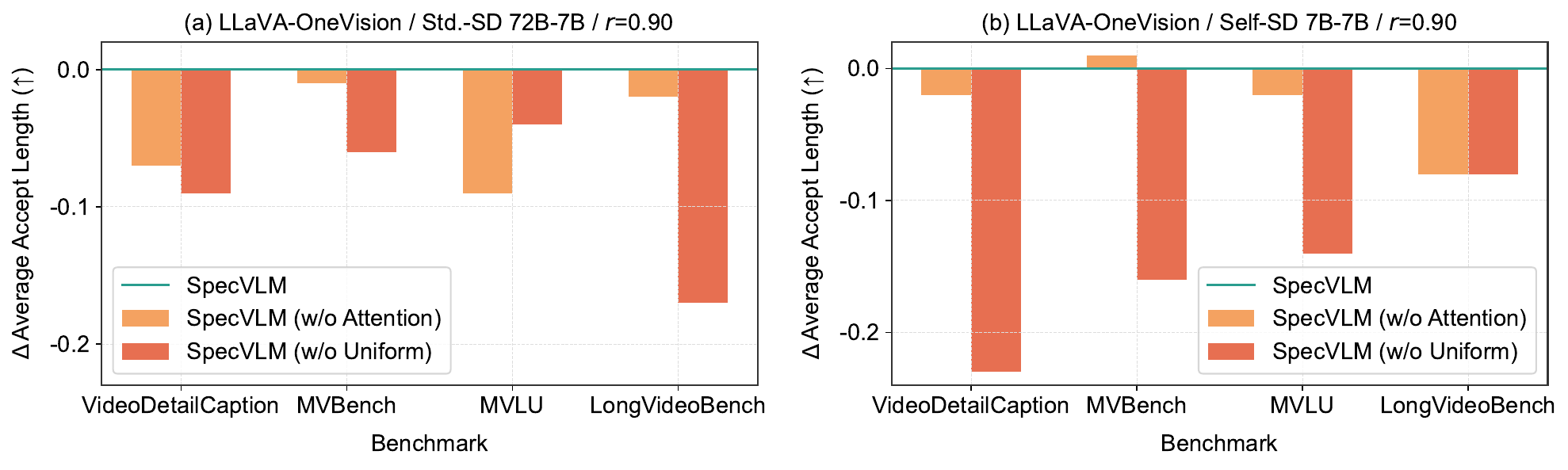}
    \caption{
    Change in average accepted length ($\Delta$) of the full \method and its ablated variants.
    }
    \label{llava_ablation}
\end{figure*}

\subsection{Main Result}
\label{main_result}
We evaluate the efficiency of different baselines in \cref{table_1,table_2}.
For LLaVA-OneVision series, \method achieves a speedup ratio up to 2.68$\times$ and 1.33$\times$ under Std.-Sd and Self-SD, respectively. For Qwen2.5-VL series, up to 2.11$\times$ and 1.59$\times$ speedup are attained accordingly. 
The default pruning ratio is set to 90\% to maximally reduce the KV cache size. Still, we include experiments on different pruning ratios in \cref{scaling_law}. 
The case study and computation breakdown of \method are elaborated in \cref{apd:case_study,appendix:breakdown}.

\paragraph{Token pruning significantly enhances the performance of speculative decoding.} Compared to vanilla autoregressive decoding, SD-Tree yields a basic speedup through the draft-and-verify process. When incorporated with video token pruning, the speedup ratio is largely boosted due to the utilization of an enhanced draft model with reduced KV cache. For Std.-SD, video token pruning improves decoding speed by 29\% and 43\% for LLaVA-OneVision-72B and Qwen2.5-VL-32B on VideoDetailCaption compared to SD-Tree. For Self-SD, video token pruning facilitates a 1.24$\times$ to 1.33$\times$ speedup for LLaVA-OneVision-7B and a 1.37$\times$ to 1.59$\times$ speedup for Qwen2.5-VL-7B. Additionally, we further note that the influence of video token pruning would become even greater as the video length increases, which can be inferred from the trend revealed in \cref{bar_and_plot} (a).

\begin{table*}[!t]
    \centering
    \footnotesize
    \setlength{\tabcolsep}{4pt} 
    \begin{tabular}{@{}l | cccccccccccc@{}}
        \toprule
        Decoding Step & 0 & 1& 2&3	&4&	5&	6&7	& 8	& 9	& Avg (First 10 Steps)&	Avg (All Steps) \\
        \midrule
        Average Accepted Length	&4.3&	3.1	&3.7	&3.4&3.7&2.95&3.6&3.6&2.5&3.1&3.39
        &3.48 \\
        \bottomrule
    \end{tabular}
    \caption{Average accepted length vs. decoding steps.}
    \label{table_decoding_steps}
\end{table*}

\paragraph{\method shows superior performance across various datasets and model architectures.}~\cref{table_1,table_2} illustrate that \method outperforms random token pruning under a high pruning ratio. On VideoDetailCaption, the average accept length $\tau$ of LLaVA-OneVision-7B using our method drops by only 5\% under 90\% video token reduction, which is 40\% of the degradation observed with SD-Rand. For Qwen2.5-VL series and on other datasets, although the sensitivity to pruning is lower, our method consistently demonstrates higher speculation accuracy to varying degrees.
Moreover, a higher value of average accept length $\tau$ in \method also implies a higher theoretical upper bound on overall speedup, demonstrating the potential of our method to be integrated with other draft model optimization techniques.

\subsection{Scaling Law for Pruning Ratio} 
\label{scaling_law}
For a comprehensive study, we conduct the experiments across a wide spectrum of pruning ratios as shown in \cref{llava_saling_comparison}, in which \method achieves a consistently higher speculation accuracy. 
Moreover, as the pruning ratio increases, the gap between \method and random token pruning gradually widens, demonstrating the effectiveness of \method under high pruning ratios.




\subsection{Ablation Study}
\label{ablation_study}

\paragraph{Necessity of Attention Guidance.}
Here, we compare our method against attention-agnostic pruning to validate the necessity of attention guidance.
We adopt a window-based approach commonly similar to StreamingLLM cache~\citep{xiaoefficient}, which retains initial and trailing language tokens along with video tokens within a fixed window. Specifically, we place the window over the front, middle, and end portions of the video tokens, retaining a fixed number of tokens according to the pruning ratio. 
\cref{llava_baseline} (a) shows that our attention-based method (SD-Attention, i.e., \method without Stage II) outperforms window-based approaches by enabling finer-grained token selection and higher speculation accuracy under a fixed KV budget.

\paragraph{Impact of Top-P Informative Token Retention.}
To verify the effectiveness of our Top-P informative token retention strategy, we adopt uniform pruning in space without the guidance of the attention signal, as depicted in \cref{llava_ablation}. In most cases, preserving highly informative tokens improves average accept length. 
For LLaVA-OneVision / Self-SD 7B-7B, an outlier appears on MVBench, which we attribute to numerous repetitive action sequences in its subset, resulting in more pronounced spatial redundancy.
\paragraph{Impact of Spatially Uniform Token Reduction.}
Similarly, we remove the second stage of our pruning strategy in~\cref{llava_ablation} to study the effect of spatially uniform token reduction. The variant of relying solely on attention signals for token retention suffers from indistinguishability issues in the uniformly low-attention area and leads to the loss of structural information and a drop in speculation accuracy, further supporting our observation in~\cref{longtail}.

\subsection{Exploration of Diverse Pruning Criteria}
Inspired by recent work~\citep{tao2024dycoke,shen2025fastvid} that explores temporal redundancy, we aim to address the question: \textit{Can pruning based on temporal redundancy offer greater benefits?} 
To this end, we compare the following baselines:

\begin{itemize}[itemsep=1.5pt, topsep=1.5pt, leftmargin=20pt]
\item[(1)] SD-Uniform: SD with uniform pruning based on token positions within the spatial layout.
\item[(2)] SD-Frame: SD with full-frame dropping at regular temporal intervals.
\item[(3)] SD-FastVID: Frame-level pruning based on Top-K similarity of consecutive frame transitions, following FastVID~\citep{shen2025fastvid}.
\item[(4)] SD-DyCoke: Token-level temporal merging adapted from DyCoke~\citep{tao2024dycoke}.
\end{itemize}

\cref{llava_baseline} (b) suggests that temporal redundancy-based methods consistently underperform compared to spatially uniform pruning (our focus)—and even random pruning—within the SD framework.
We believe this indicates that the draft model benefits from retaining the overall temporal structure and distribution of camera shots to achieve better alignment, while some spatial information is more redundant in this context. 

\subsection{Impact of Decoding Steps}
\label{generation_length}
To validate the effectiveness of our method in selectively preserving visual information to enhance speculation accuracy, we performed experiments on the average acceptance length vs. decoding steps using LLaVA-OneVision-72B/7B with a 90\% pruning ratio setting on VideoDetailCaption. We averaged the performance over 50 instances for evaluation. The results are presented in \cref{table_decoding_steps}, from which we can conclude: 
(i) \textbf{The average accept length of the initial steps does not significantly differ from the average across all steps.} The average for the first 10 decoding steps (3.39) is not notably lower than the overall average (3.48), and there is no significant upward trend within the first 10 steps. We attribute this to the fact that each decoding step benefits from the visual information retained during the prefill stage, allowing for a relatively high speculation accuracy from the outset. 
(ii) \textbf{The first decoding step, in fact, exhibits a significantly higher average accepted length.} This is likely because the beginning of the generation often follows a fixed sentence structure, making it easier to speculate.

\section{Related Work}
\label{related_work}
\subsection{Speculative Decoding}
\paragraph{Speculative Decoding for LLMs.}
Speculative decoding is shown to be an effective approach to accelerate LLMs while maintaining the original output distribution. It relies on two key processes: efficient drafting and parallel verification. Initial explorations~\citep{leviathan2023fast,guo2023longcoder,kim2023speculative,xia2023speculative} attempt to use existing small LLMs as draft models to ensure reliable speculation. Self-speculative methods~\citep{xia2025swift,zhang2024draft,song2025knnsd} use partial layers of the original model to generate predictions, without introducing extra models. Recent advancements~\citep{cai2024medusa,li2024eagle,du2024glide} focus on enhancing the efficiency of the drafting stage. These works attain high acceleration ratios through a specially trained draft model with reduced latency. Meanwhile, tree-based speculation methods~\citep{li2024eagle2,li2024eagle,miao2023specinfer} are proposed to boost the average accept length, by predicting multiple candidates and forming draft trees. Among the aforementioned draft models, the success of the previous state-of-the-art EAGLE method~\citep{li2024eagle2} highlights the \textit{autoregressive structure} as a key factor in improving the accuracy of the speculation. However, autoregressive draft models need to maintain their own KV caches, which introduces additional memory overhead when faced with long video input.

\paragraph{Long-Context Speculative Decoding.}
Long-context speculative decoding methodologies~\citep{sun2024triforce,chen2024magicdec,yang2025longspec} are specially developed to mitigate the substantial overhead of KV cache. Specifically, TriForce~\citep{sun2024triforce} introduces a hierarchical speculation to tackle the two bottlenecks of model weights and KV cache. MagicDec~\citep{chen2024magicdec} reassesses the trade-off between throughput and latency. LongSpec~\citep{yang2025longspec} proposes a memory-efficient draft model with a constant memory footprint. 
\cref{appendix:long-context sd} elaborates on the reasons why these methods are not suitable for Vid-LLMs.

\subsection{Visual Token Reduction}
Compared to information-dense text, visual tokens often exhibit high redundancy, making token reduction an effective way to reduce computational and memory overhead. Recent studies largely focus on training-free visual token reduction methods based on token importance and redundancy. FastV~\citep{chen2024image} selects important visual tokens after layer 2 using attention scores of MLLMs. While SparseVLM~\citep{zhang2024sparsevlm} evaluates the visual relevance of text tokens and performs pruning based on the attention scores of a subset of text tokens. VisionZip~\citep{yang2024visionzip} reduces visual redundancy in the vision encoders. DART~\citep{wen2025stop} prunes tokens based on its duplication with other tokens. Currently, video token reduction draws increasing attention due to the high volume of video tokens. DyCoke~\citep{tao2024dycoke} performs token merging across frames and reduces KV cache dynamically. PrunVID~\citep{huang2024prunevid} identifies static and dynamic tokens across frames, and selectively prunes visual features relevant to question tokens. FastVID~\citep{shen2025fastvid} partitions frames into segments and introduces a density-based token pruning strategy. VidCom$^2$~\citep{liu2025video} dynamically adjusts compression intensity based on frame uniqueness.
Notably, recent~\citet{tang2025adaptive} applies temporal processing during video sampling by performing key frame extraction, whereas ours prunes visual tokens post-encoder, making the two methods orthogonal.
\section{Conclusion}
\label{conclusion}

We propose \method, the first \textit{training-free} speculative decoding framework tailored for accelerating video LLMs. 
Building on the low speculation sensitivity to token pruning, \method leverages verifier-guided attention to remove redundant video tokens, significantly reducing the draft model’s KV cache without compromising generation quality.
\method achieves up to 2.68$\times$ speedup on LLaVA-OneVision-72B and 2.11$\times$ on Qwen2.5-VL-32B across multiple video understanding tasks.
We hope our work inspires further research on latency-efficient Vid-LLM reasoning from a token sparsity perspective and believe \method will serve as a valuable tool for the community to enable efficient video comprehension.


\section{Acknowledgements}
The work was supported by the Major Research Program of Zhejiang Provincial Natural Science Foundation (No.~LD24F020015), Zhejiang Province "Leading Talent of Technological Innovation Program" (No. 2023R5214), Beijing Natural Science Foundation(Grant No.L233034), and Fundamental Research Funds for the Beijing University of Posts and Telecommunications (Grant No.2025TSQY01).

\clearpage
\section{Ethical Considerations}
All experiments in this work are conducted using open-source datasets and models. Our research focuses solely on improving inference efficiency and does not involve any sensitive data, human subjects, or commercial use~\footnote{\url{https://huggingface.co/datasets/MLVU/MVLU} (CC-BY-NC-SA-4.0 license)
}
\footnote{\url{https://huggingface.co/datasets/OpenGVLab/MVBench} (CC-BY-4.0 license)
}
\footnote{\url{http://github.com/longvideobench/LongVideoBench} (CC-BY-NC-SA-4.0 license)
}
\footnote{\url{https://huggingface.co/datasets/lmms-lab/VideoDetailCaption} (CC-BY-4.0 license)
}.

\section{Limitation}
While our enhanced speculative decoding framework brings clear benefits for accelerating Vid-LLMs, there are a few limitations to consider. First, our method is primarily applicable to resource-constrained long-video scenarios, where memory bandwidth becomes the dominant bottleneck. Second, we introduce an additional draft model during inference. Although its computational overhead is relatively small compared to the target model (or can be avoided using the Self-SD setting), the choice of the draft model requires careful consideration to achieve optimal speedup. Moreover, to avoid introducing training as an additional variable, we use an existing Vid-LLM as the draft model without further fine-tuning. This design choice imposes certain constraints on the maximum achievable acceleration.
Nevertheless, our method has the potential to be seamlessly integrated with smaller and faster draft models, as it only requires a one-time pruning of the draft model's KV cache during the prefilling stage. As the technology for training lightweight Vid-LLMs is still in its early stages and lacks suitable candidates, we believe future work can explore more draft model optimization techniques~\citep{li2024eagle,du2024glide} to address this limitation.

\bibliography{custom}
\clearpage

\appendix

\section{Experimental Details}
\label{appendix:implementation Details}
\paragraph{Task and Benchamark Details.}
Since \method mainly focuses on the lossless acceleration during the decoding stage, we select video captioning and video description as our tasks, which require the model to summarize and describe the video by understanding detailed subjects and actions, and involve generating
relatively long text paragraphs. 
Unlike conventional video question answering (VQA), we prompt the model to generate a description rich in visual information instead of direct answer selection. 
We sample videos from mainstream video understanding benchmarks, including VideoDetailCaption~\citep{vdc}, MVBench~\citep{li2024mvbench}, MVLU~\citep{zhou2024mlvu}, and LongVideoBench~\citep{wu2024longvideobench}, to ensure a comprehensive coverage of different durations and varying scenarios. For each benchmark, 50 instances are randomly sampled. For LLaVA-OneVision-72B and LLaVA-OneVision-7B, we uniformly sample 64 and 128 frames to generate a 196 $\times$ 64 and 196 $\times$ 128 video token input, respectively. For Qwen2.5-VL series, we adjust the FPS to generate input of comparable length.

\paragraph{Implementation Details.}
All experiments are implemented on 8 NVIDIA A100 GPUs. We utilize the default attention implementation of LLaVA-OneVision (scaled-dot-product-attention~\citep{vaswani2017attention}), and output the attention scores using its Python implementation when required. 
Given each query, the target model is employed to generate 256 tokens following greedy decoding. When video tokens are pruned, we remove the corresponding video features based on the pruning ratio $r$. 
During evaluation, the default pruning ratio $r$ is set to 90\%, based on the low sensitivity property validated in \cref{section_3.2}. The hyperparameter $\lambda_r$ is determined at the model level through a single offline evaluation on a small calibration set consisting of 2–4 randomly sampled instances per task (with no overlap with the test set). In \cref{table_1,table_2}, we set $\lambda_r$ to 0.4 and 0.5, respectively.

\section{Impact of Tree Attention} \label{Appendix:Tree}
\begin{figure} [t!] 
\centering
    \includegraphics[scale=0.3]{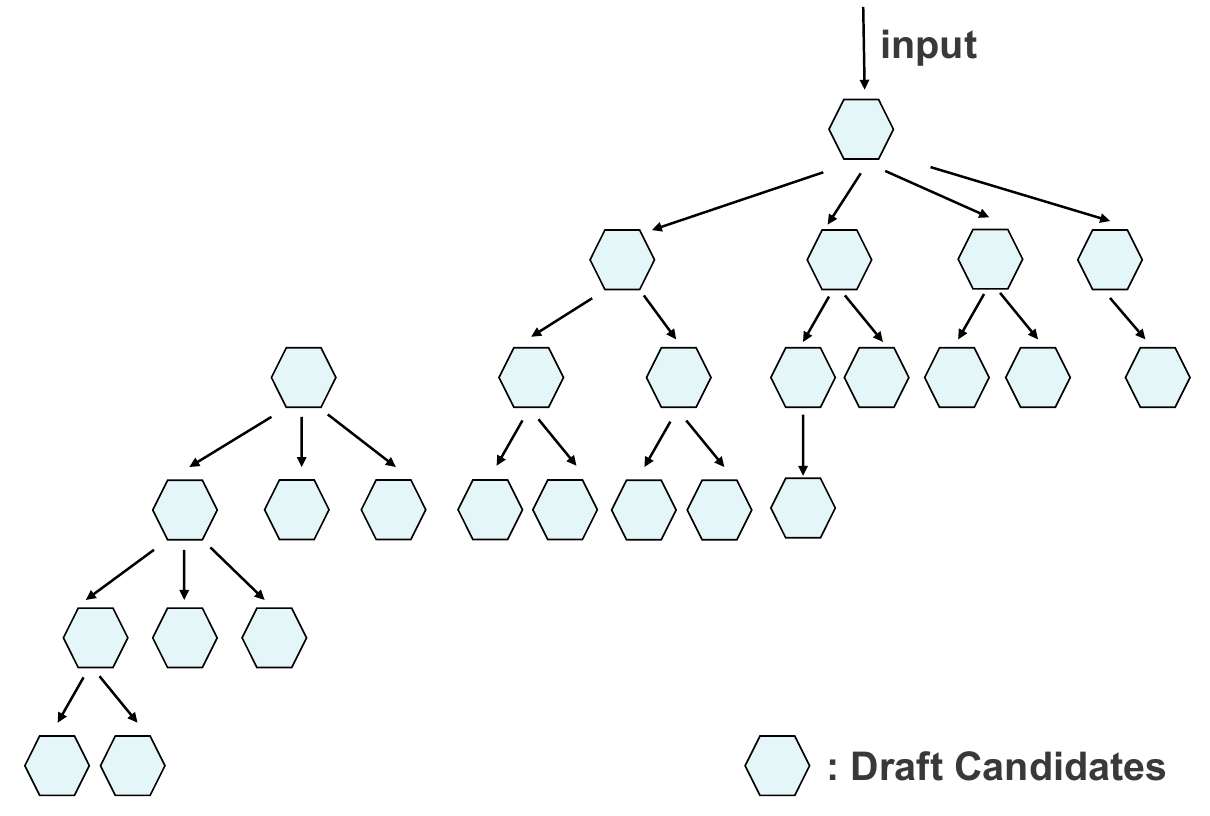}
    \caption{
    The draft tree structure of \method. 
    }
    \label{draft_tree_structure}
\end{figure}

As illustrated in \cref{table_draft_tree}, using a tree structure for drafting and verification greatly enhances the average accept length by validating multiple candidates in a single forward. When incorporated with tree attention, \method achieves a higher speculation accuracy. The draft tree structure adopted in \method is depicted as \cref{draft_tree_structure}. 
The chosen tree structure is motivated by the intuition that candidate tokens with higher probabilities merit deeper and wider expansion, whereas low-probability tokens should not be further explored.

\begin{table}
    \centering
    \footnotesize
  \setlength{\tabcolsep}{4pt} 
    \begin{tabular}{@{}l|c@{}}
        \toprule
        Method & $\tau$ \\
        \midrule
        SD-Chain & 3.12 \\
        SD-Tree & 3.68  \\
        \midrule
        \method-Chain & 2.79  \\
        \method & 3.48  \\
    
        \bottomrule
    \end{tabular}
    \caption{Average accept length $\tau$. ``*-Chain'' refers to a sequential drafting process without a tree structure. Results are tested using LLaVA-OneVision-72B / 7B on VideoDetailCaption. By default, $r=90\%$. }
    \label{table_draft_tree}
\end{table}

\begin{table}[!t]
    \centering
    \footnotesize
  \setlength{\tabcolsep}{4pt} 
    \begin{tabular}{@{}lcc@{}}
        \toprule
        Operation & Vanilla & \method \\
        \midrule
        Target Model Prefilling & 24.01 & 24.01 \\
        Target Model Decoding & 87.03 & 32.47 \\
        Draft Model Prefilling & - & 0.82 \\
        Video Token Pruning & - & 0.06 \\
        \midrule
        Latency & 111.04 & 57.36 \\
        \bottomrule
    \end{tabular}
    \caption{Inference time breakdown (s) of LLaVA-OneVision-72B / 7B. Output length is set to 256.}
    \label{table_breakdown}
\end{table}

\section{Breakdown of Computation}
\label{appendix:breakdown}
Apart from accelerating the decoding stage of target Vid-LLMs, \method introduces minimal overhead in the pruning process, as illustrated in \cref{table_breakdown}. Owing to video token pruning, the prefill length of the draft model is substantially reduced, and the additional prefilling time becomes negligible compared to target model inference time. In this work, we significantly reduce the target model’s decoding time—a level of efficiency that cannot be achieved by prior methods relying solely on token reduction. 

\section{Why is Long-context SD Technologies Not Suitable for Vid-LLMs?}
\label{appendix:long-context sd}
As mentioned earlier in~\cref{related_work}, long-context SD techniques are designed to address the KV cache bottleneck of conventional SD, combined with KV cache sparsification techniques~\citep{feng2025adakvoptimizingkvcache,feng2025identifycriticalkvcache,xiaoefficient}. To this end, they employ StreamingLLM caches \citep{xiaoefficient} or sliding window caches for the draft model to manage the KV cache budget. However, the lack of awareness of the visual modality prevents these methods from effectively operating on video tokens. On the one hand, video tokens exhibit substantial redundancy, with a large portion being repetitive or highly similar—unlike discrete language tokens. On the other hand, there exists a distinct attention pattern between language and video tokens, as illustrated in \cref{attention_map}, making modality-aware methods better suited to exploit this property. If StreamingLLM-style caches or sliding windows are applied naively, the draft model can only attend to a small portion of the uncompressed visual content, thereby failing to capture the overall semantics of the video.

\section{Discussion on Smaller Draft Model}\label{apd:small draft model} 
In this section, we discuss the effectiveness of \method in a smaller draft model scenario. Although the minimum draft model used in our experiment is 7b, we believe our method still has benefits for smaller draft models from the following two perspectives. (i) \textbf{SpecVLM prunes redundant video tokens at inference time, reducing the training cost and architectural complexity of high-quality small draft models for video understanding.} Existing small draft models for LLMs like EAGLE do not support long-context input sequences. Training small draft models to match the context length of Vid-LLMs would require massive computation, and it remains questionable whether these small models can achieve the same accuracy at video input context lengths~\citep{sun2024triforce}. Therefore, reducing the input length of the small draft model through our video token pruning strategy allows it to perform efficient speculation without the need for costly long-context training. 
(ii) \textbf{For smaller draft models, the overhead caused by input length still persists.} To simulate a scenario similar to EAGLE, we tested the latency of a single layer of the LLaVA-OneVision-0.5B model at varying input lengths. Latency is calculated on average of 100 decoded tokens on a single Nvidia A100 GPU. The results in \cref{small_draft_latency} show that the latency of the small draft model increases significantly with input length, eventually becoming multiple times larger than the original. Given the increasing trend of video inputs (e.g., Long video input with million-level video tokens~\citep{chen2024longvila} ), the additional draft overhead from the KV Cache would gradually undermine the lightweight nature of smaller draft models. In other words, combining draft models with token pruning still provides a performance benefit.
In future work, we will investigate memory- and data-efficient training strategies for Vid-LLMs~\citep{li2024eagle,hu2022lora,zhang2025train,zhang2025sf}, with the goal of constructing more optimized draft models that remain effective in memory-constrained devices~\cite{zhou2025floe} and scalable to multi-tenant  settings~\cite{zhang2025hmi}.

\begin{table}
    \centering
    \footnotesize
    \setlength{\tabcolsep}{2pt} 
    \begin{tabular}{@{}l|cccccc@{}}
        \toprule
        Input Sequence Length (K) & 1 & 4 & 8 & 16 & 32 & 48 \\
        \midrule
        Draft Latency (ms) & 1.0 & 1.16 & 1.46 & 2.08 & 3.29 & 4.53 \\
        \bottomrule
    \end{tabular}
    \caption{Layer latency variation of smaller draft model with growing input sequence length.}
    \label{small_draft_latency}
\end{table}

\section{Case Study}\label{apd:case_study}
Two illustrative examples are presented in~\cref{appendix:case1,appendix:case2}. 
The target model performs verification using the original video tokens, ensuring lossless input. Meanwhile, the draft model relies on a 90\% pruned version of the video tokens, allowing for efficient yet accurate speculation by retaining essential visual cues and maintaining a coherent understanding of the overall video structure.

\begin{figure*} [t!] 
\centering
    \includegraphics[scale=0.75]{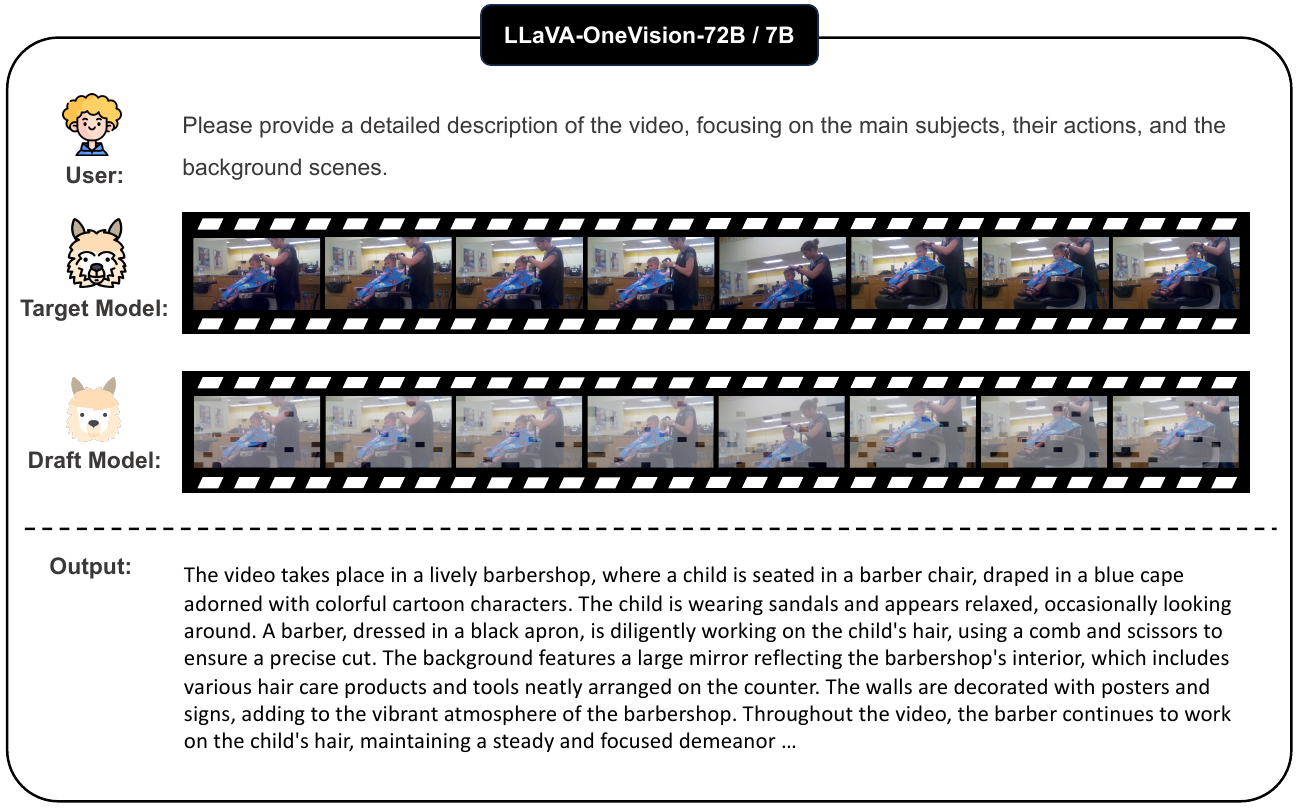}
    \caption{
    Visualization of \method on VideoDetailCaption using LLaVA-OneVision-72B / 7B.
    }
    \label{appendix:case1}
\end{figure*}

\begin{figure*} [t!] 
\centering
    \includegraphics[scale=0.75]{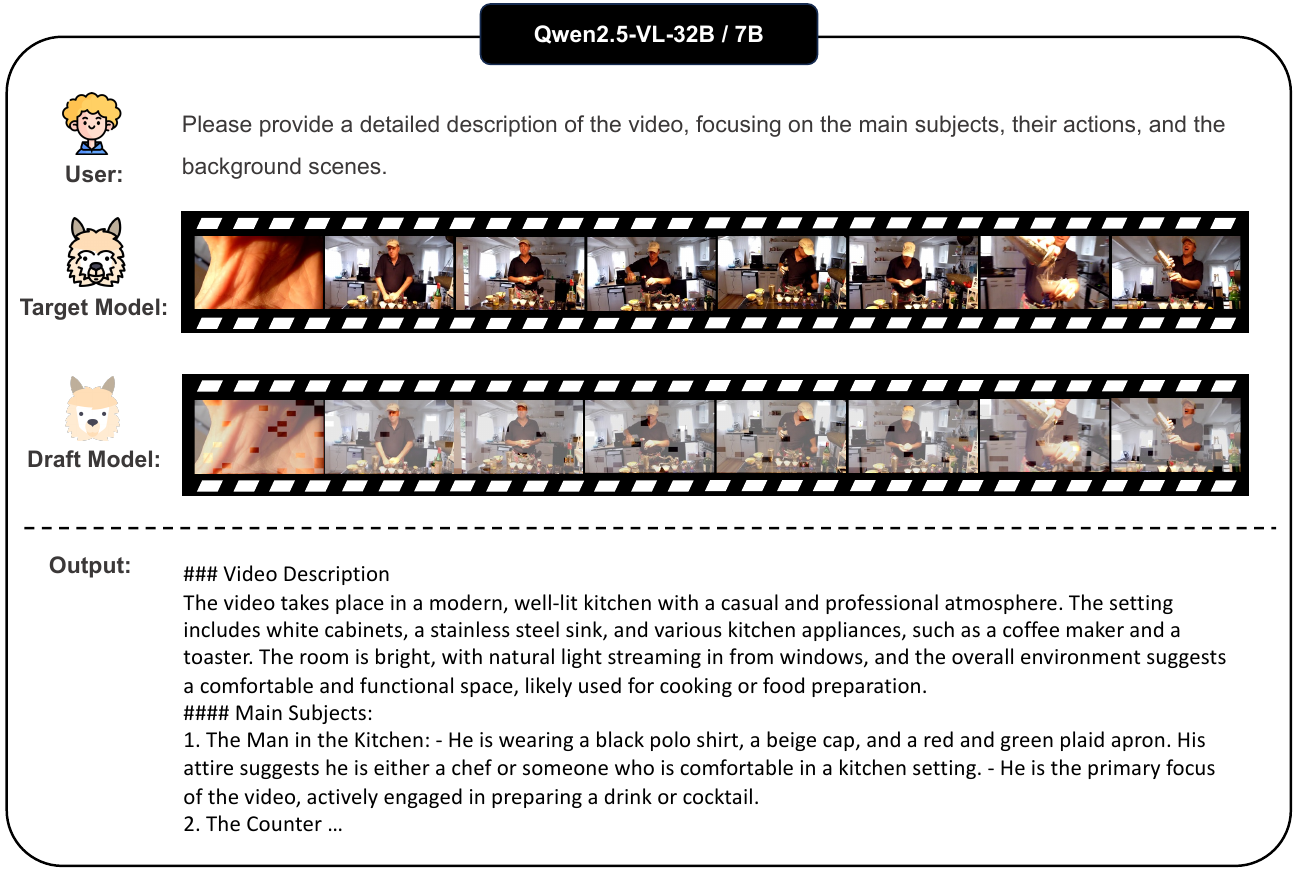}
    \caption{
    Visualization of \method on VideoDetailCaption using Qwen2.5-VL-32B / 7B.
    }
    \label{appendix:case2}
\end{figure*}

\end{document}